\documentclass[runningheads]{llncs}

\usepackage{eccv}

\usepackage{eccvabbrv}
\usepackage{graphicx}
\usepackage{booktabs}
\usepackage{makecell}
\usepackage{multirow}
\usepackage{multicol}
\usepackage{rotating}
\usepackage{wrapfig}
\usepackage{xcolor}
\usepackage{colortbl}
\usepackage[accsupp]{axessibility}  % Improves PDF readability for those with disabilities.

\definecolor{LightBlue}{rgb}{0.8235,0.9737,0.9882}
\definecolor{LightGray}{HTML}{ececec}
\definecolor{LightGreen}{HTML}{effdee} % e4fde1
\definecolor{LightYellow}{HTML}{fffee0} % fffec8, fffdaf
\definecolor{LightOrange}{HTML}{fee7da} % fee2d2
\definecolor{textGreen}{HTML}{008000}
\definecolor{textRed}{HTML}{cc0000}
\newcommand{\red}[1]{\textcolor{red}{#1}}
\newcommand{\blue}[1]{\textcolor{blue}{#1}}

\usepackage{hyperref}

\usepackage{orcidlink}

\newcommand{\myparagraph}[1]{\vspace{3pt}\noindent{\bf #1}}

\newcommand{\ours}{LRSLAM}

\begin{document}

\title{LRSLAM: Low-rank Representation of Signed Distance Fields in Dense Visual SLAM System} 

\titlerunning{LRSLAM}
%\titlerunning{LRSLAM: Low-rank Representation of Signed Distance Fields in Dense Visual SLAM System}

\author{Hongbeen Park\inst{1}\orcidlink{0009-0003-2633-288X} \and
Minjeong Park\inst{2}\orcidlink{0009-0008-0075-6234} \and
Giljoo Nam\inst{3}\orcidlink{0000-0002-1822-1501} \and
Jinkyu Kim\inst{1}\orcidlink{0000-0001-6520-2074}}

\authorrunning{H. Park et al.}

\institute{Dept of Computer Science and Engineering, Korea University, Seoul, Korea \and
Dept of Electrical and Electronic Engineering, Yonsei University, Seoul, Korea \and
Meta Reality Labs, Pittsburgh, PA 15222, USA  \\
% Springer Heidelberg, Tiergartenstr.~17, 69121 Heidelberg, Germany
%\email{\{qkrghdqls1, jinkyu.kim\}@korea.ac.kr}
}

\maketitle

\def\thefootnote{*}\footnotetext{Corresponding author: J. Kim (jinkyukim@korea.ac.kr)}

\begin{abstract}
  Simultaneous Localization and Mapping (SLAM) has been crucial across various domains, including autonomous driving, mobile robotics, and mixed reality. Dense visual SLAM, leveraging RGB-D camera systems, offers advantages but faces challenges in achieving real-time performance, robustness, and scalability for large-scale scenes. Recent approaches utilizing neural implicit scene representations show promise but suffer from high computational costs and memory requirements. ESLAM introduced a plane-based tensor decomposition but still struggled with memory growth. Addressing these challenges, we propose a more efficient visual SLAM model, called LRSLAM, utilizing low-rank tensor decomposition methods. Our approach, leveraging the Six-axis and CP decompositions, achieves better convergence rates, memory efficiency, and reconstruction/localization quality than existing state-of-the-art approaches. Evaluation across diverse indoor RGB-D datasets demonstrates LRSLAM's superior performance in terms of parameter efficiency, processing time, and accuracy, retaining reconstruction and localization quality. Our code will be publicly available upon publication.

\end{abstract}

\section{Introduction}\label{sec:intro}
\begin{figure*}[t]
    \centering
    \includegraphics[width=\linewidth]{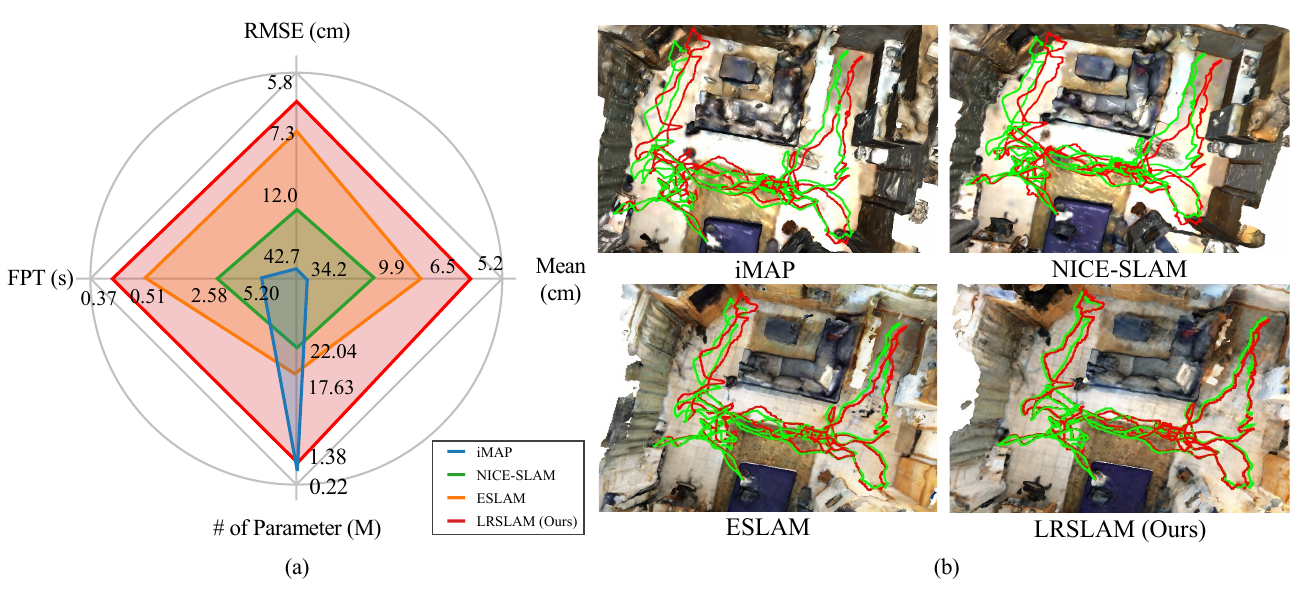}

    \caption{(a) \textbf{Comparison with SOTA Approaches.} Our model is clearly positioned as an efficient yet effective dense visual SLAM method. Our proposed LRSLAM requires fewer parameters, faster frame processing time (FPT), and better reconstruction accuracy (regarding ATE mean and RMSE). This is further validated by our (b) \textbf{Qualitative Comparison of Scene Reconstruction and Localization} between our proposed LRSLAM and the state-of-the-art approaches, including iMAP~\cite{imap}, NICE-SLAM~\cite{nice} and ESLAM~\cite{eslam}. Ours shows comparable or better reconstruction and localization accuracy with highly compact representations.}
    \label{fig:teaser}
    \vspace{-25pt}
\end{figure*}

Simultaneous Localization and Mapping (SLAM) has been an essential technology in various domains, such as autonomous driving~\cite{bresson2017simultaneous, chen2019suma++}, indoor/outdoor mobile robotics~\cite{dworakowski2021robot, ruan2021smart, liu2020novel}, and mixed reality~\cite{covolan2020mapping, jinyu2019survey, singandhupe2019review}. Recently, dense visual SLAM approaches based on an RGB-D camera system with additional depth information have been actively explored due to the advantages of simple sensor configuration. Despite promising results, their high computational costs make it challenging to achieve (i) real-time performance, (ii) robustness, and (iii) scalability to deal with large-scale scenes~\cite{tosi2024nerfs,imap,nice,eslam}. These are crucial factors in rendering a SLAM system truly effective for real-world applications. 

Learning underlying representations of scene geometry and appearance is pivotal in building such a visual SLAM system. With remarkable success with Neural Radiance Fields (NeRF) techniques~\cite{nerf}, recent work~\cite{imap, nice} suggests that neural implicit scene representation can be utilized to learn geometry and appearance representations, optimizing a 3D map and camera poses for a visual SLAM system. Yet, their cubic memory growth rate necessitates employing voxel grids with reduced resolutions, sacrificing intricate geometric details. More recently, ESLAM~\cite{eslam} leverages plane-based tensor decomposition to achieve efficient and accurate localization and reconstruction. Despite its promising outcomes, it still has a quadratic memory growth rate, which is still challenging for a real-time visual SLAM system.  

Following this stream of visual SLAM models, we propose a more efficient model with a linear memory growth rate, thereby improving both the efficiency and accuracy of SLAM tasks, i.e., localization and reconstruction. To this end, we focus on compactly factorizing the 3D geometry and appearance of a scene into parameterized low-rank components (more compact than ESLAM's plane-based representation), enabling a compact yet expressive scene representation. Specifically, we propose a new tensor decomposition method, called Six-axis decomposition, which factorizes the three planes in the tri-plane representation into six axis-aligned feature tensors, thus holding an efficient memory complexity of O($n$). Also, we propose a hybrid scene representation using both the conventional CP decomposition~\cite{carroll1970analysis} and our new Six-axis decomposition to further improve the overall performance of SLAM.

In summary, we propose an efficient visual SLAM method, called LRSLAM, which leverages a combination of low-rank tensor decomposition methods (i.e., our proposed Six-axis decomposition and CP decomposition) to provide a better convergence rate, memory efficiency, and reconstruction/localization quality. We observe that a hybrid use of these two tensor decomposition methods provides a notable benefit in the following two aspects: (1) CP decomposition allows compact and fast encoding of the geometry features (once converged, it also helps to learn the appearance features) and (2) Six-axis (SA) decomposition allows to learn detailed appearance features with more expressive yet efficient decomposition, which is essential for the tracking task. We conduct thorough evaluations across diverse indoor RGB-D datasets, including ScanNet~\cite{scannet}, TUM RGB-D~\cite{tum}, and Replica~\cite{replica}. In our experiments, our model uses remarkably fewer parameters (87.3\%--90.1\% fewer than ESLAM~\cite{eslam}) and shows faster processing time (4.3\%--73.2\% than ESLAM~\cite{eslam}), retaining reconstruction and localization accuracy. 

We summarize our contributions as follows:
\begin{itemize}
    \item We propose a novel tensor decomposition method called Six-axis decomposition. This method compactly factorizes the tri-planes into six axis-aligned feature tensors with a linear memory growth rate.\vspace{0.5em}
    \item We leverage the novel Six-axis decomposition together with the traditional CP decomposition to achieve compact yet effective RGB-D SLAM performance.\vspace{0.5em}
    \item Our extensive experiments with three public datasets (i.e., ScanNet, TUM RGB-D, and Replica) validate the effectiveness of our proposed approach, which significantly reduces the demand for parameters while showing a faster convergence rate with matched or outperforming reconstruction and localization performance. 
\end{itemize}

\section{Related Work}

\myparagraph{Visual SLAM.} Visual SLAM techniques can mainly be categorized into three types depending on the data source: (i) Visual-only SLAM~\cite{covolan2020mapping, taketomi2017visual, fuentes2015visual, yousif2015overview}, which utilizes a mono (or multiple) camera system and thus needs to accurately estimate depth from cameras only, which is still technically challenging. (ii) Visual-inertial SLAM~\cite{gui2015review, huang2019visual}, which relies on additional inertial measurement units (IMUs) to improve the overall accuracy. However, their system is prone to noise in calibration and inertial measurements. (iii) RGB-D SLAM~\cite{imap, nice, eslam, zhang2021survey} utilizes additional depth information and thus provides reliable and improved performance. Recent approaches have significantly improved with RGB-D SLAM, reporting better accuracy and robustness, albeit with drawbacks such as increased memory and power requirements. In this paper, we follow the stream of RGB-D SLAM and aim to improve its accuracy, satisfying low memory and power requirements.

\myparagraph{Neural Scene Representation for Visual SLAM.}
Neural Radiance Fields (NeRF, \cite{nerf}) have had a significant impact on various applications such as large-scale 3D reconstructions by utilizing neural implicit representations. Recent work suggests that such neural implicit representations can be applied to dense visual SLAM systems, showing promising localization performance by accurately representing the scene. A landmark work is iMAP~\cite{imap}, which utilizes an implicit neural scene representation in a real-time SLAM system using an RGB-D camera, proving its capability to optimize a 3D map and camera poses. However, its performance is often limited due to the model's capacity for encoding a wide scene. To solve this, NICE-SLAM~\cite{nice} extends iMAP by representing the scene with voxel grid features and converting them into occupancies using pre-trained MLPs. However, their model's cubic memory growth rate leads to the use of low-resolution voxel grids and the loss of fine geometric details. Recently, ESLAM~\cite{eslam} has been introduced for efficient yet accurate localization and reconstruction by leveraging plane-based tensor decomposition. Despite their promising results, their model still exhibits quadratic memory growth, leading to constraints and sub-optimal performance for real-time visual SLAM systems. Thus, in this work, we propose a more efficient model with a linear memory growth rate, further enhancing the efficiency and accuracy of localization and reconstruction. 

\myparagraph{Concurrent Work using 3D Gaussian Splatting.} 3D Gaussian Splatting (3DGS)~\cite{kerbl20233d} is gaining increasing attention as a new scene representation. 3DGS is well-known for its faithful 3D reconstruction quality and the efficient rendering paradigm. As concurrent work of ours, several SLAM systems are proposed to utilize 3DGS as a scene representation~\cite{li2024sgs,yan2023gs,yugay2023gaussian,matsuki2023gaussian,keetha2023splatam,huang2023photo}. However, as discussed in \cite{li2024sgs,yan2023gs}, the large memory consumption is also an issue in 3DGS-based SLAM systems, because they often fail to manage the number of Gaussians in unseen areas~\cite{tosi2024nerfs}.

% \section{Tensor Decompositions for NeRF-based SLAM}
% \section{Six-axis Decomposition for Efficient Scene Representation}
\section{Efficient Scene Representations}
\label{sec:prelim}

\begin{figure*}[t]
    \centering
    \includegraphics[width=\linewidth]{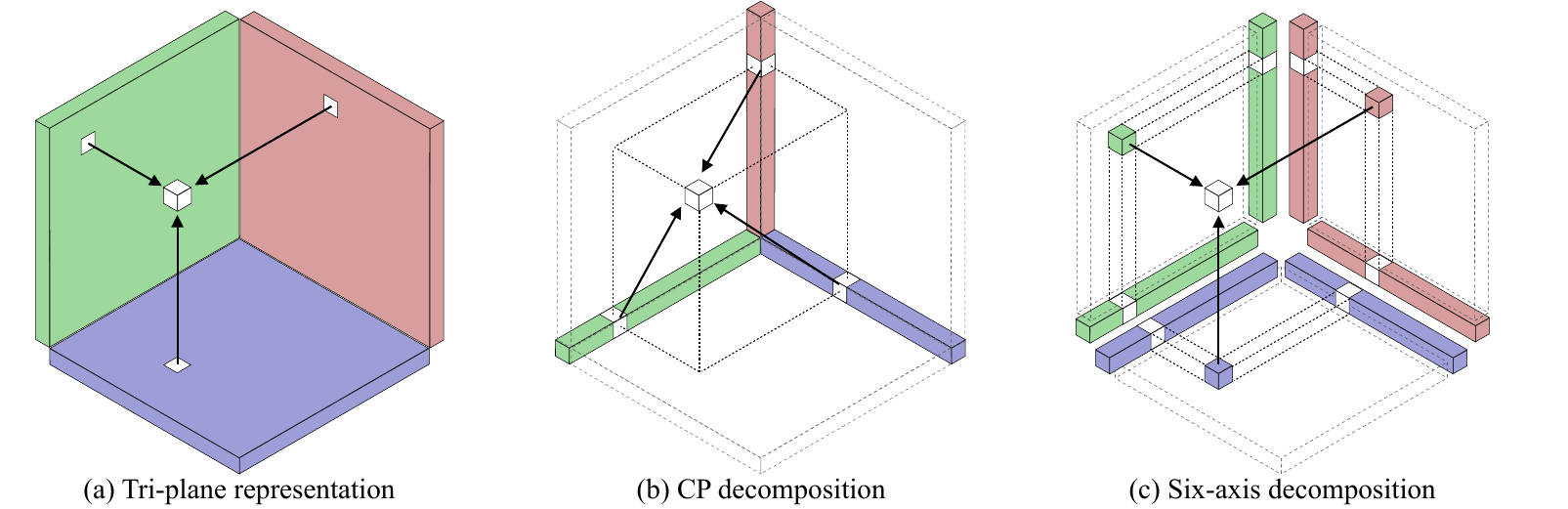}
    \caption{(a) \textbf{Tri-plane representation} factorizes a 4D tensor into three feature planes (Eq.~\ref{eq:triplane}). (b) \textbf{CP decomposition} factorizes a 4D tensor into a sum of outer products of three axis-aligned low-rank feature tensors (Eq.~\ref{eq:cp}). (c) Our proposed \textbf{Six-axis decomposition} factorizes a 4D tensor into a set of six axis-aligned low-rank feature tensors (Eq.~\ref{eq:sixaxis}).}
    \label{fig:comparison-of-decomposiion}
    \vspace{-2em}
\end{figure*}

The 3D geometry of a scene can be depicted using a signed distance field (SDF), denoted as $f_{SDF}: \mathbb{R}^3 \rightarrow \mathbb{R}$. This function maps a 3D location $\mathbf{p}$ to a scalar value $s$, which is the distance from $\mathbf{p}$ to the nearest surface in the scene.
This function can be efficiently implemented using a combination of factorized tensors and a small neural network~\cite{chan2022efficient,chen2022tensorf}. In formal terms, $f_{SDF} = \mathtt{MLP}(f(\mathbf{p}))$, where $f: \mathbb{R}^3 \rightarrow \mathbb{R}^C$ is a function that takes a 3D location $\mathbf{p}$ as input and outputs a length-$C$ feature vector, and $\mathtt{MLP}: \mathbb{R}^C \rightarrow \mathbb{R}$ is a multilayer perceptron that decodes this feature vector into a scalar value $s$.
The compactness and accuracy of these representations depend on the specific method employed for the function $f$. In this section, we discuss two preliminary methods, i.e., tri-plane representation and CP decomposition and then introduce our novel tensor decomposition method, the Six-axis (SA) decomposition.

\myparagraph{Tri-plane Representation.}
The tri-plane representation~\cite{chan2022efficient} employs three 2D feature planes, i.e., $f_{xy},f_{yz},f_{zx}\in\mathbb{R}^{L\times L\times C}$ each with a spatial resolution of $L\times L$ and $C$ feature channels. 
To query the feature vector at a 3D location $\mathbf{p}$, we first project $\mathbf{p}$ onto each axis-aligned plane and aggregate the three retrieved features from respective planes. 
Chan et al.~\cite{chan2022efficient} suggested that summation can serve as an efficient feature aggregation method, yielding the tri-plane representation as follows:
\begin{equation}
  \label{eq:triplane}
  f_\text{tri-plane}(\mathbf{p}) = {f_{xy}(\mathbf{p})+f_{yz}(\mathbf{p})+f_{zx}(\mathbf{p})}.
\end{equation}
This representation has O$(n^2)$ space complexity where $n$ is the side length of the scene. 
See Fig.~\ref{fig:comparison-of-decomposiion} (a).

\myparagraph{CP Decomposition.}
CP decomposition provides a more compact representation of $f()$ than the tri-plane representation. It factorizes a 4D tensor into a sum of outer products of three axis-aligned rank-one tensors as follows:
\begin{align}
  \label{eq:cp}
  f_\text{CP}(\mathbf{p}) = \{\sum_{i=1}^{k} f_x^{(i)} \otimes f_y^{(i)} \otimes f_z^{(i)}\}(\mathbf{p})
\end{align}
where $f_x^{(i)}, f_y^{(i)}, f_z^{(i)}\in\mathbb{R}^{L\times C}$ are factorized low-rank tensors of three modes for the $i$-th component. 
As illustrated in Fig.~\ref{fig:comparison-of-decomposiion} (b), the CP decomposition has O$(n)$ space complexity and can provide the most compact representation of a scene. 
However, as discussed in \cite{chen2022tensorf}, CP decomposition may cause information loss due to is extreme compactness. 

\myparagraph{Six-axis Decomposition.}
Here we present our novel Six-axis (SA) decomposition. The key idea is to factorize the three planes in the tri-plane representation into a set of six axis-aligned low-rank feature tensors.  
Given three feature planes of rank $k$, i.e., $f_{xy},f_{yz},f_{zx}$, they are further factorized as the sum of outer products of $k$ rank one tensors: e.g., $f_{xy}=\sum_{i=1}^{k}f_{x_y}^{(i)}\otimes f_{y_x}^{(i)}$ where $f_{x_y}^{(i)}, f_{y_x}^{(i)} \in\mathbb{R}^{L\times C}$ are axis-aligned low-rank tensors. Similarly, other feature planes, $f_{yz}$ and $f_{zx}$, can be decomposed into the sum of outer products of $k$ axis-aligned low-rank tensors, yielding the Six-axis decomposition as follows:
\begin{equation}
  \label{eq:sixaxis}
  f_\text{SA}(\mathbf{p}) = \{\sum_{i=1}^{k} f_{x_y}^{(i)} \otimes f_{y_x}^{(i)}\}(\mathbf{p}) + \{\sum_{i=1}^{k} f_{y_z}^{(i)} \otimes f_{z_y}^{(i)}\}(\mathbf{p}) + \{\sum_{i=1}^{k} f_{z_x}^{(i)} \otimes f_{x_z}^{(i)}\}(\mathbf{p})
\end{equation}
where $f_{y_z}^{(i)}, f_{z_y}^{(i)}, f_{z_x}^{(i)}, f_{x_z}^{(i)}\in\mathbb{R}^{L\times C}$ are axis-aligned features. 
As shown in Fig.~\ref{fig:comparison-of-decomposiion} (c), compared to the tri-plane representation, Six-axis decomposition has advantages in holding the efficient memory complexity of O($n$), retaining the high capability of scene encoding and decoding. 

%%%%%%%%%%%%%%%

\section{Low-rank Representations for RGB-D SLAM}
\begin{figure*}[t]
    \centering
    \includegraphics[width=\linewidth]{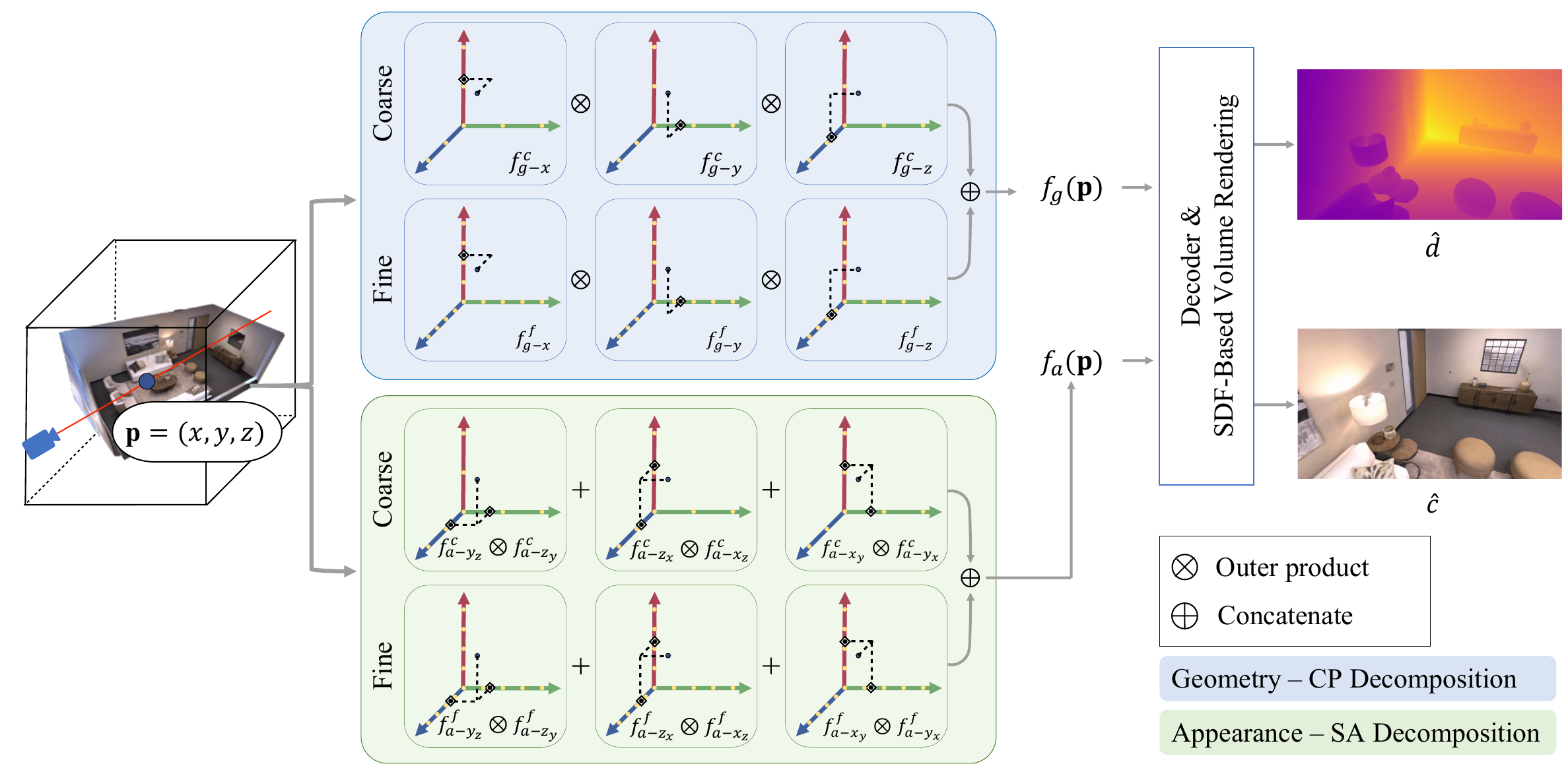}
    \caption{An overview of our proposed model, called LRSLAM. Our model utilizes a combination of low-rank tensor decomposition methods to provide a better convergence rate, memory efficiency, and reconstruction quality. Specifically, we utilize the CP decomposition to represent the geometry of a scene (see top, $f_\text{g}$) and use our Six-axis decomposition for reconstructing color (see bottom, $f_\text{a}$).}
    \label{fig:main}
\end{figure*}

In this section, we propose LRSLAM, a memory-efficient RGB-D SLAM that employs low-rank representations of signed distance fields. 
We first provide a summary of the baseline approach ESLAM, which is the state-of-the-art RGB-D SLAM method. We then delve into the specifics of employing the novel SA decomposition together with the traditional CP decomposition to achieve optimal RGB-D SLAM performance.

\subsection{Baseline Method}
We use ESLAM~\cite{eslam} as the baseline approach for the RGB-D SLAM system. 
Given a sequence of RGB-D frames $\{I_i, D_i\}^{M}_{i=1}$, ESLAM jointly estimates camera poses $\{R_i | t_i\}^{M}_{i=1}$, an SDF $f_{SDF}: \mathbb{R}^3 \rightarrow \mathbb{R}$, and its appearance $f_{RGB}: \mathbb{R}^3 \rightarrow \mathbb{R}^3$. 

\myparagraph{Scene Representation.}
The main technical contribution of ESLAM is the employment of tri-plane representation to efficiently model the SDF of a scene. 
Specifically, ESLAM models a scene with twelve axis-aligned multi-resolution feature planes, six for geometry and the other six for appearance. 
Formally, it has three coarse geometry feature planes $\{F^{c}_{g\text-xy},F^{c}_{g\text-xz},F^{c}_{g\text-yz}\}$ and three fine geometry feature planes $\{F^{f}_{g\text-xy},F^{c}_{g\text-xz},F^{f}_{g\text-yz}\}$. 
Appearance feature planes are similarly defined as $\{F^{c}_{a\text-xy},F^{c}_{a\text-xz},F^{c}_{a\text-yz}\}$ and $\{F^{f}_{a\text-xy},F^{f}_{a\text-xz},F^{f}_{a\text-yz}\}$. 
To query the feature vector of a 3D location $\mathbf{p}$, they sum the bilinearly interpolated features of each plane and then concatenate the coarse and fine features together. For example, a geometric feature at $\mathbf{p}$ is obtained:
\begin{equation}
\begin{aligned}
\label{eq:eslam_feature}
  & f^{c}_{g}(\mathbf{p}) = F^{c}_{g\text-xy}(\mathbf{p}) + F^{c}_{g\text-xz}(\mathbf{p}) + F^{c}_{g\text-yz}(\mathbf{p}) \\
  & f^{f}_{g}(\mathbf{p}) = F^{f}_{g\text-xy}(\mathbf{p}) + F^{f}_{g\text-xz}(\mathbf{p}) + F^{f}_{g\text-yz}(\mathbf{p}) \\
  & f_{g}(\mathbf{p}) = [ f^{c}_{g}(\mathbf{p}) ; f^{f}_{g}(\mathbf{p}) ].
\end{aligned}
\end{equation}
An appearance feature $f^{f}_{a}(\mathbf{p})$ can be obtained in the same manner. 
The features are then decoded into the final SDF value $\phi_g(\mathbf{p})$ and color $\phi_a(\mathbf{p})$:
\begin{equation}
\begin{aligned}
\label{eq:eslam_decoder}
  & \phi_g(\mathbf{p}) = \mathtt{MLP}_g(f_{g}(\mathbf{p})) \\
  & \phi_a(\mathbf{p}) = \mathtt{MLP}_a(f_{a}(\mathbf{p})).
\end{aligned}
\end{equation}
In practice, truncated signed distance field (TSDF) is used because SLAM systems only require geometric information around the surfaces in a scene. 
The TSDF value $\phi_g(\mathbf{p})$ and its color $\phi_a(\mathbf{p})$ can effectively model a 3D scene. 

\myparagraph{SDF-based Volume Rendering.}
Given the current camera pose estimates, random pixels are chosen and the corresponding rays are obtained. 
For each ray, $N$ points $\{\mathbf{p}\}_{i=1}^{N}$ are sampled and their corresponding TSDF $\phi_g(\mathbf{p}_i)$ and raw color $\phi_c(\mathbf{p}_i)$ are computed using Eq.~\ref{eq:eslam_decoder}.
Specifically, $N$ points are composed of $N_s$ points from stratified sampling and $N_t$ points from importance sampling. 
For volume rendering, TSDF values are converted into volume densities via a mapping function $\phi_{d}(\mathbf{p}_i)=\beta \cdot \textrm{Sigmoid}(-\beta \cdot \phi_g(\mathbf{p}_i))$, where $\beta$ is a learnable parameter to control the thickness of surface boundaries. The color $\hat{\mathbf{c}}$ and depth $\hat{d}$ of the ray can be computed as follows:
\begin{equation}
\begin{aligned}
\label{eq:volume_render}
    & \hat{c} = \sum_{i=1}^{N}{w_i\phi_a(\mathbf{p}_i)} \textrm{,} \quad \hat{d} = \sum_{i=1}^{N}{w_iz_i} \\
    & w_i=\textrm{exp}(-\sum_{j=1}^{i-1}\phi_{d}(\mathbf{p}_j))(1-\textrm{exp}(-\phi_{d}(\mathbf{p}_i)))
\end{aligned}
\end{equation}
where $z_i$ is the depth of a given point $\mathbf{p}_i$. Please refer to \cite{eslam} for more details about the pixel and point sampling strategies.

\myparagraph{Simultaneous Localization and Mapping.}
Using a set of generated rays, ESLAM optimizes camera poses and scene parameters by minimizing a loss function that consists of several terms: depth loss, color loss, free-space loss, and TSDF loss.
The depth loss minimize the difference between rendered depth $\hat{d}$ and sensor-measured depth. The color losses works similarly for the rendered color $\hat{\mathbf{c}}$. 
The free-space loss prevents empty spaces form having valid SDF values, and the TSDF loss makes the SDF values $\phi_g(\mathbf{p})$ confirm to the sensor-measured depth. 
We refer to \cite{eslam} for more implementation details about the loss functions and overall SLAM pipeline. 

\subsection{\ours}
We propose to use low-rank representations of 4D tensors to enhance the efficiency of RGB-D SLAM. Our approach involves the use of CP decomposition to depict the geometric aspect of a scene via a truncated signed distance field (TSDF). At the same time, we employ our SA decomposition to represent the scene's appearance. In the following, we provide a detailed explanation of our proposal and discuss the reasoning behind our choice of this hybrid representation. All the experiments shown in this paper base on the ESLAM~\cite{eslam} framework, but with a different scene representation. 

\myparagraph{CP Decomposition for Geometry.}
We use a set of low-rank feature tensors to represent the geometry of a scene with the CP decomposition. Formally, the geometry feature vector $f_\text{g}(\mathbf{p}) \in \mathbb{R}^{C}$ at a point $\mathbf{p}$, is computed as the sum of outer products of three axis-aligned vectors as follows:

\begin{equation}
\begin{aligned}
\label{eq:lr_geo}
  & f^c_g(\mathbf{p}) = \{\sum_{i=1}^{k_g} f_{g\text-x}^{c(i)} \otimes {f_{g\text-y}^{c(i)}} \otimes {f_{g\text-z}^{c(i)}}\}(\mathbf{p}) \\
  & f^f_g(\mathbf{p}) = \{\sum_{i=1}^{k_g} f_{g\text-x}^{f(i)} \otimes {f_{g\text-y}^{f(i)}} \otimes {f_{g\text-z}^{f(i)}}\}(\mathbf{p}) \\
  & f_{g}(\mathbf{p}) = [ f^{c}_{g}(\mathbf{p}) ; f^{f}_{g}(\mathbf{p}) ],
\end{aligned}
\end{equation}
where 
$f_{g\text-x}^{c(i)}$ is the $i$-th coarse-level rank-one tensor for geometry. Other features are similarly defined. The final TSDF value is decoded via a small MLP:
\begin{equation}
  \phi_g(\mathbf{p}) = \mathtt{MLP}_g(f_{g}(\mathbf{p})).
\end{equation}

\myparagraph{SA Decomposition for Appearance.}
For appearance, we use our SA decomposition.
The appearance feature vector $f_\text{a}(\mathbf{p}) \in \mathbb{R}^{C}$ at a point $\mathbf{p}$ is computed as the sum of outer products of six axis-aligned vectors:
\begin{align}
  & f^c_a(\mathbf{p}) = \{\sum_{i=1}^{k_a} f_{a\text-x_y}^{c(i)} \otimes f_{a\text-y_x}^{c(i)}\}(\mathbf{p}) + \{\sum_{i=1}^{k_a} f_{a\text-y_z}^{c(i)} \otimes f_{a\text-z_y}^{c(i)}\}(\mathbf{p}) + \{\sum_{i=1}^{k_a} f_{a\text-z_x}^{c(i)} \otimes f_{a\text-x_z}^{c(i)}\}(\mathbf{p})  \label{eq:clr} \\
  & f^f_a(\mathbf{p}) = \{\sum_{i=1}^{k_a} f_{a\text-x_y}^{f(i)} \otimes f_{a\text-y_x}^{f(i)}\}(\mathbf{p}) + \{\sum_{i=1}^{k_a} f_{a\text-y_z}^{f(i)} \otimes f_{a\text-z_y}^{f(i)}\}(\mathbf{p}) + \{\sum_{i=1}^{k_a} f_{a\text-z_x}^{f(i)} \otimes f_{a\text-x_z}^{f(i)}\}(\mathbf{p})  \label{eq:clr} \\
  & f_{a}(\mathbf{p}) = [ f^{c}_{a}(\mathbf{p}) ; f^{f}_{a}(\mathbf{p}) ],
\end{align}
where $f_{a\text-x_y}^{c(i)}$ is the $i$-th coarse-level rank-one tensor for appearance. Remaining features can be similarly defined. 
The final color is decoded via a small MLP:
\begin{equation}
  \phi_a(\mathbf{p}) = \mathtt{MLP}_a(f_{a}(\mathbf{p})).
\end{equation}

\myparagraph{Rationale Behind the Hybrid Representation.}
As shown in Sec.~\ref{sec:prelim}, both CP and SA decomposition has a significant advantage in space complexity, which is our main motivation of using such low-rank representations in SLAM. While the SA decomposition has better representational power than CP decomposition, we empirically find that using the hybrid representation, i.e., CP for geometry and SA for appearance, yields better performance in RGB-D SLAM, as shown in Fig.~\ref{fig:ablation}. 
We identify two key reasons for this. 
Firstly, the geometry of a scene typically contains lower frequency information than its appearance, making the CP decomposition sufficient for representing geometries in many cases.
Second, during optimization, CP decomposition converges faster than SA decomposition due to its simplicity. We find that the early convergence of geometry helps the appearance optimization because of the geometric dependency of color volume rendering in Eq.~\ref{eq:volume_render}.
Fig.~\ref{fig:ablation} shows the ablation study that supports our choice of hybrid representation.

\section{Experiments}

We evaluate the effectiveness of our novel compact decomposition method for dense visual SLAM systems on various publicly available real and synthetic datasets. We also conduct a detailed ablation study to support the feasibility of our design choice in terms of speed and accuracy. 

\subsection{Experimental Setup}

\myparagraph{Datasets.}
We conduct our experiments on the following three widely-used dense visual SLAM benchmarks: Replica~\cite{replica}, ScanNet~\cite{scannet}, and TUM RGB-D~\cite{tum}. We evaluate the localization performance, i.e., camera tracking errors, for all the three datasets. In addition, we evaluate the reconstruction performance using the Replica dataset which provides the ground-truth geometries. We preprocess the datasets in the same way used in recent work~\cite{nice, eslam}.

\myparagraph{Implementation Details.}\label{sec:implementation}
In our experiments, we use two-layer MLPs (with 64 input channels and 16 hidden layer channels) for our decoders. For SDF-based volume rendering on a small-scale Replica~\cite{replica}, we set $N_s=32$ and $N_t=8$, which are sampled by stratified and importance sampling, respectively. We use $N_s=48$ and $N_t=8$, which are sampled similarly for the other datasets. Our coarse axis-aligned feature tensors use a resolution of 24 cm for both geometry and appearance, while the fine feature tensors use 6 cm and 3 cm for geometry and appearance, respectively. All axis-aligned features are set to have 32-dimensional tensors. Lastly, we set $k_g=2$ for CP decomposition and $k_a=16$ for SA decomposition. We provide more implementation details in the supplemental material.

\myparagraph{Evaluation Metrics.}
We follow recent work~\cite{nice,eslam} for evaluation metrics. For the scene geometry evaluation, we use both 2D and 3D metrics, where we use the L1 loss on depth maps from ground truth and reconstructed meshes across 1000 randomly sampled camera poses. Further, for 3D metrics, we use reconstruction accuracy (in cm), reconstruction completion (in cm), and completion ratio. For a fair comparison, the volume resolution is set to 1cm, and we remove unseen regions that are not visible from any camera frustum. In addition, we use the Absolute Trajectory Error (ATE, \cite{tum}) RMSE and Mean to evaluate localization. By default, we run five independent runs and report the average results unless otherwise stated.

\begin{wraptable}[22]{r}{0.75\textwidth}
    \vspace{-1.2cm}
    \caption{\textbf{Computation and Memory Efficiency.} We compare runtime and memory efficiency between other state-of-the-art approaches, including NICE-SLAM~\cite{nice} and ESLAM~\cite{eslam}. We measure the frame processing time (FPT), number of parameters for scene geometry and appearance feature planes. We assume a spatial resolution of $L\times L\times L$.}
    \centering
    \setlength{\tabcolsep}{2pt}
    \renewcommand{\arraystretch}{1.3} 
    \resizebox{0.75\textwidth}{!}{%
        \begin{tabular}{lllcccrl}
        \toprule
          \multicolumn{2}{c}{\multirow{2}{*}{Data}} & \multirow{2}{*}{Method} & \multirow{2}{*}{FPT(s)} & & \# of Parameters & & \multirow{2}{*}{Complexity}\\
          \cmidrule{5-7}
          & & & & $f_{g}$ & $f_{a}$ & Total & \\
       \midrule
        
        \multicolumn{2}{c}{\multirow{3}{*}{\rotatebox[origin=c]{90}{\small Replica}}}  & NICE-SLAM~\cite{nice} & 2.27 & 6.42M & 5.70M & 12.18M & O$(L^3)$ \\
        & & ESLAM~\cite{eslam} & 0.23 & 1.40M & 5.38M & 6.79M & O$(L^2)$ \\
        & & 
        \cellcolor{LightGray} LRSLAM (ours) &
        \cellcolor{LightGray} 0.22 &
        \cellcolor{LightGray} 0.03M &
        \cellcolor{LightGray} 0.83M &
        \cellcolor{LightGray} 0.86M &
        \cellcolor{LightGray} O$(L)$ \\
        & & & 
        \cellcolor{LightYellow} \red{(4.3\%$\downarrow$)} &
        \cellcolor{LightYellow} \red{(97.9\%$\downarrow$)} &
        \cellcolor{LightYellow} \red{(84.6\%$\downarrow$)} &
        \cellcolor{LightYellow} \red{(87.3\%$\downarrow$)} &
        \cellcolor{LightYellow} \\
        \midrule
        
        \multicolumn{2}{c}{\multirow{3}{*}{\rotatebox[origin=c]{90}{\small ScanNet}}} & NICE-SLAM~\cite{nice} & 2.58 & 11.63M & 10.36M & 22.04M & O$(L^3)$\\
        & & ESLAM~\cite{eslam} & 0.51 & 3.66M & 13.98M & 17.63M & O$(L^2)$\\
        & & 
        \cellcolor{LightGray} LRSLAM (ours) &
        \cellcolor{LightGray} 0.37 &
        \cellcolor{LightGray} 0.05M &
        \cellcolor{LightGray} 1.33M &
        \cellcolor{LightGray} 1.38M &
        \cellcolor{LightGray} O$(L)$ \\
        & & &
        \cellcolor{LightYellow} \red{(27.5\%$\downarrow$)} &
        \cellcolor{LightYellow} \red{(98.6\%$\downarrow$)} &
        \cellcolor{LightYellow} \red{(90.5\%$\downarrow$)} &
        \cellcolor{LightYellow} \red{(90.1\%$\downarrow$)} & 
        \cellcolor{LightYellow}\\
        \midrule
        
        {\multirow{3}{*}{\rotatebox[origin=c]{90}{\small TUM}}} & {\multirow{3}{*}{\rotatebox[origin=c]{90}{\small RBG-D}}} & NICE-SLAM~\cite{nice} & 13.21 & 23.56M & 21.02M & 44.64M & O$(L^3)$\\
        & & ESLAM~\cite{eslam} & 4.56 & 1.40M & 5.36M & 6.77 M & O$(L^2)$\\
         & &  
         \cellcolor{LightGray} LRSLAM (ours) & 
         \cellcolor{LightGray} 1.22 & 
         \cellcolor{LightGray} 0.03M & 
         \cellcolor{LightGray} 0.81M & 
         \cellcolor{LightGray} 0.84M & 
         \cellcolor{LightGray} O$(L)$ \\
         & & & 
         \cellcolor{LightYellow} \red{(73.2\%$\downarrow$)} & 
         \cellcolor{LightYellow} \red{(97.9\%$\downarrow$)} & 
         \cellcolor{LightYellow} \red{(84.9\%$\downarrow$)} & 
         \cellcolor{LightYellow} \red{(87.6\%$\downarrow$)} & 
         \cellcolor{LightYellow}\\
        \bottomrule
    \end{tabular}
    }
    \label{table:score_llff} 
\end{wraptable}

%Analysis of runtime and memory efficiency of our method with recent representation in terms of average frame processing time (FPT), number of parameters, and model space complexity \wrt scene side-length $L$.

\subsection{Analysis of Computation and Memory Efficiency}
We start by evaluating the computation cost and memory efficiency. As we summarize in Table~\ref{table:score_llff}, we report the average frame processing time (FPT), the number of parameters for scene geometry and appearance feature planes, and memory complexity regarding big-O notation. We compare ours with other state-of-the-art approaches, such as NICE-SLAM~\cite{nice} and ESLAM~\cite{eslam}. Note that we report scores for a scene \texttt{room0} of Replica~\cite{replica}, \texttt{scene0000} of ScanNet~\cite{scannet}, and \texttt{fr1/desk} in TUM RGB-D~\cite{tum} datasets. We use a single NVIDIA A100 GPU to measure such scores. As expected, our proposed LRSLAM uses a remarkably reduced number of parameters (87.3\%--90.1\% fewer parameters than ESLAM), while its processing time becomes faster (4.3\%--73.2\%) than ESLAM. Such gains become more apparent with real-world scenes (i.e., ScanNet and TUM RGB-D) than synthetic scenes (i.e., Replica).

\subsection{Evaluation of Mapping and Localization Performance}

\begin{figure*}[!t]
    \centering
    \includegraphics[width=\linewidth]{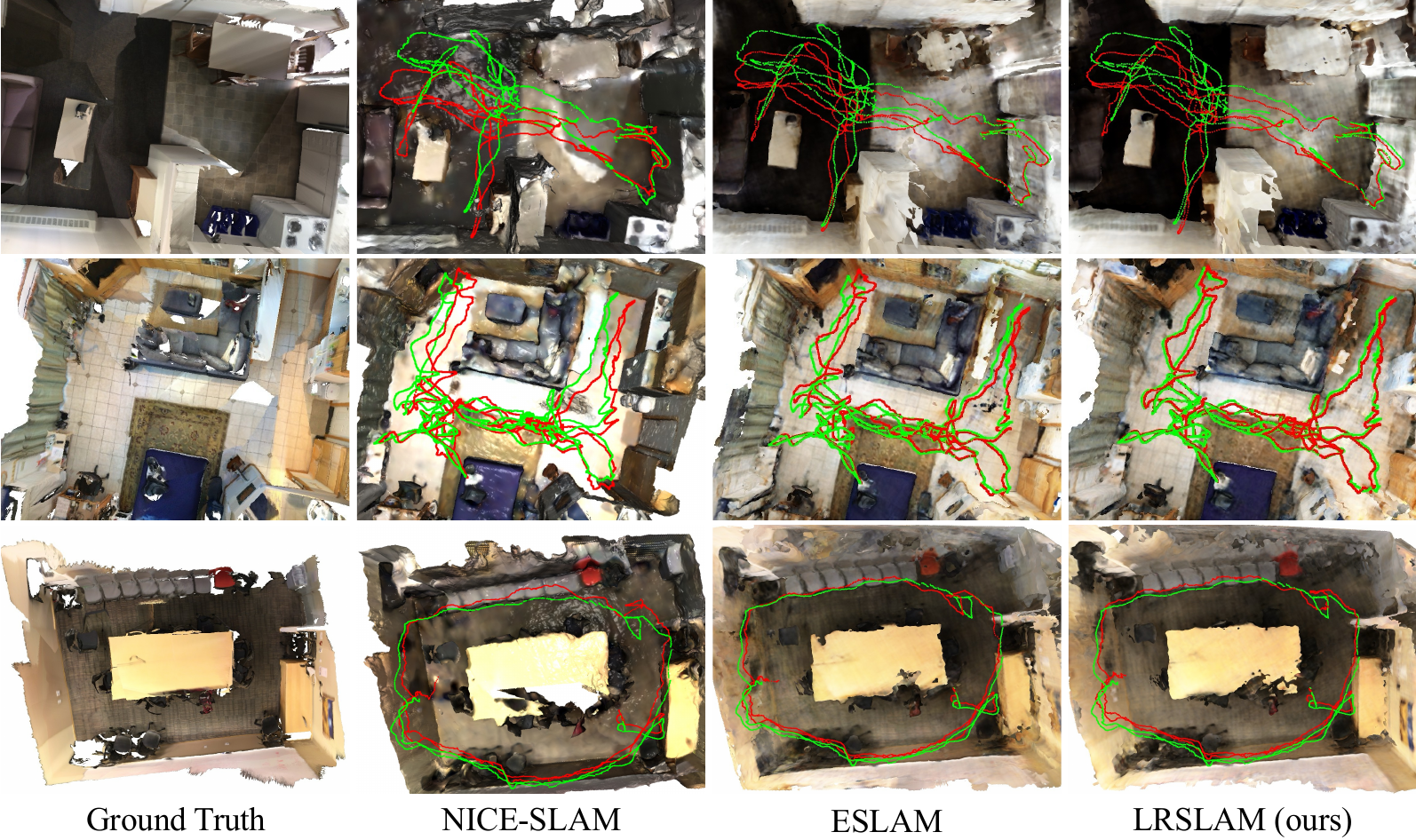}
    \caption{\textbf{Reconstruction and Localization on ScanNet~\cite{scannet}.} We qualitatively compare the quality of 3D reconstruction and localization between our LRSLAM and existing approaches, including NICE-SLAM~\cite{nice} and ESLAM~\cite{eslam}. For localization, the green trajectory is from the ground truth, and the red is the tracking results from each method. Despite using ten times more compact representation, our approach provides matched reconstruction and localization performance.}
    \label{fig:scannet}
\end{figure*}

%\label{sec:scannet-localization}
{
\setlength{\tabcolsep}{3pt}
\renewcommand{\arraystretch}{1.3} 
\begin{table*}[!t]
    \caption{\textbf{Localization Results on ScanNet~\cite{scannet}.} We compare with existing approaches, including NICE-SLAM~\cite{nice} and ESLAM~\cite{eslam} in terms of ATE Mean and ATE RMSE. Our proposed method shows generally better results than the others.}
    %\vspace{-.5em}
        \resizebox{\linewidth}{!}{
        \begin{tabular}%{l@{\hspace{3mm}}l@{\hspace{2mm}}@{\hspace{2mm}}c@{\hspace{3mm}}c@{\hspace{3mm}}c@{\hspace{3mm}}c@{\hspace{3mm}}c@{\hspace{3mm}}c@{\hspace{2mm}}@{\hspace{2mm}}c@{\hspace{2mm}}}
        {llrrrrrrr}
            \toprule
            Method & ATE & Sc. 0000 & Sc. 0059 & Sc. 0106 & Sc. 0169 & Sc. 0181 & Sc. 0207 & Avg.\\
            \midrule
            NICE-SLAM\cite{nice} & Mean$\downarrow$ & 9.9 $\pm$ 0.4 & 11.9 $\pm$ 1.8 & 7.0 $\pm$ 0.2 & 9.2 $\pm$ 1.0 & 12.2 $\pm$ 0.3 & 5.5 $\pm$ 0.3 & 9.3 $\pm$ 0.7 \\
            & RMSE$\downarrow$ & 12.0 $\pm$ 0.5 & 14.0 $\pm$ 1.8 & 7.9 $\pm$ 0.2 & 10.9 $\pm$ 1.1 & 13.4 $\pm$ 0.3 & 6.2 $\pm$ 0.4 & 10.7 $\pm$ 0.7 \\
            % \hline
            % NICER-SLAM\cite{nicer} & Mean & \small \phantom{0}9.9 $\pm$ \phantom{0}0.4 & \small 11.9 $\pm$ 1.8 & \small \phantom{0}7.0 $\pm$ 0.2 & \small \phantom{0}9.2 $\pm$ \phantom{0}1.0 & \small 12.2 $\pm$ 0.3 & \small \phantom{0}5.5 $\pm$ 0.3 & \small \phantom{0}9.3 $\pm$ 0.7 \\
            %     & RMSE & \small 12.0 $\pm$ \phantom{0}0.5 & \small 14.0 $\pm$ 1.8 & \small \phantom{0}7.9 $\pm$ 0.2 & \small 10.9 $\pm$ \phantom{0}1.1 & \small 13.4 $\pm$ 0.3 & \small \phantom{0}6.2 $\pm$ 0.4 & \small 10.7 $\pm$ 0.7 \\
            \midrule
            ESLAM\cite{eslam} & Mean$\downarrow$ & 6.5 $\pm$ 0.1 & 6.4 $\pm$ 0.4 & 6.7 $\pm$ 0.1 & 5.9 $\pm$ 0.1 &  8.3 $\pm$ 0.2 & 5.4 $\pm$ 0.1 & 6.5 $\pm$ 0.2 \\
            & RMSE$\downarrow$ & 7.3 $\pm$ 0.2 & 8.5 $\pm$ 0.5 & 7.5 $\pm$ 0.1 & 6.5 $\pm$ 0.1 & 9.0 $\pm$ 0.2 & 5.7 $\pm$ 0.1 & 7.4 $\pm$ 0.2 \\

            \midrule
            \cellcolor{LightGray} LRSLAM (ours) &
            \cellcolor{LightGray} Mean$\downarrow$ &
            \cellcolor{LightGray} 5.2 $\pm$ 0.2 &
            \cellcolor{LightGray} 6.1 $\pm$ 0.1 &
            \cellcolor{LightGray} 6.7 $\pm$ 0.1 &
            \cellcolor{LightGray} 5.6 $\pm$ 0.1 &
            \cellcolor{LightGray} 7.6 $\pm$ 0.1 &
            \cellcolor{LightGray} 5.2 $\pm$ 0.1 &
            \cellcolor{LightGray} 6.1 $\pm$ 0.1 \\
            & & 
            \cellcolor{LightYellow} \red{(20.0\%$\downarrow$)}& 
            \cellcolor{LightYellow} \red{(4.7\%$\downarrow$)} & 
            \cellcolor{LightYellow} \red{(0.0\%$\downarrow$)}& 
            \cellcolor{LightYellow} \red{(5.1\%$\downarrow$)}& 
            \cellcolor{LightYellow} \red{(8.4\%$\downarrow$)}& 
            \cellcolor{LightYellow} \red{(3.7\%$\downarrow$)}& 
            \cellcolor{LightYellow} \red{(6.2\%$\downarrow$)}\\
            \cellcolor{LightGray} & 
            \cellcolor{LightGray}RMSE$\downarrow$ & 
            \cellcolor{LightGray} 5.8 $\pm$ 0.4 & 
            \cellcolor{LightGray} 8.2 $\pm$ 0.1 & 
            \cellcolor{LightGray} 7.6 $\pm$ 0.1 & 
            \cellcolor{LightGray} 6.5 $\pm$ 0.1 & 
            \cellcolor{LightGray} 8.4 $\pm$ 0.1 & 
            \cellcolor{LightGray} 5.6 $\pm$ 0.1 & 
            \cellcolor{LightGray} 7.0 $\pm$ 0.2 \\
            & & 
            \cellcolor{LightYellow} \red{(20.5\%$\downarrow$)}& 
            \cellcolor{LightYellow} \red{(3.5\%$\downarrow$)} & 
            \cellcolor{LightYellow} \blue{(1.3\%$\uparrow$)}& 
            \cellcolor{LightYellow} \red{(0.0\%$\downarrow$)}& 
            \cellcolor{LightYellow} \red{(6.7\%$\downarrow$)}& 
            \cellcolor{LightYellow} \red{(1.8\%$\downarrow$)}& 
            \cellcolor{LightYellow} \red{(5.4\%$\downarrow$)}\\
            \bottomrule
        \end{tabular}
        }
    %\vspace{-1ex}
    \label{table:quantitative_scannet_localization}
    % \vspace{-1.0ex}
\end{table*}
}
%Quantitative comparision of our proposed LRSLAM with existing NeRF-based dense visual SLAM models on the ScanNet dataset\cite{scannet} for localization accuracy.

\myparagraph{Evaluation on ScanNet~\cite{scannet}.}
We compare reconstruction and localization accuracy on large real scenes from ScanNet~\cite{scannet} dataset with existing state-of-the-art approaches, including NICE-SLAM~\cite{nice} and ESLAM~\cite{eslam}. In Fig.~\ref{fig:scannet}, we provide a qualitative analysis of camera localization and geometry reconstruction. Compared to the ground truth trajectory (green lines), the tracking results from our method are comparable or better performance without showing any large drifting, which confirms that our model can reconstruct precise geometry and detailed appearance with a ten times compact representation. Further, in Table~\ref{table:quantitative_scannet_localization}, we provide quantitative analysis of localization on the same dataset in terms of ATE Mean and ATE RMSE for six large real scenes. We run five times for each method, reporting the average and standard deviation. Note that we only report localization results as the dataset's ground truth meshes are incomplete. Our quantitative analysis also validates that our proposed LRSLAM generally outperforms the other state-of-the-art methods in localization, with a much lower standard deviation (thus, more stable). This confirms that our compact representation can provide significant gains in computations and memory efficiency without any degradation in localization and reconstruction (in fact, ours achieves better localization).

\begin{wraptable}[13]{r}{0.5\textwidth}
    \vspace{-1.2cm}
    \setlength{\tabcolsep}{4pt}
    \renewcommand{\arraystretch}{1.3} 
    \caption{\textbf{Localization Results on TUM RGB-D~\cite{tum}.} We compare with existing approaches, i.e., NICE-SLAM~\cite{nice} and ESLAM~\cite{eslam} in terms of ATE RMSE. All methods generally show reasonable performance, while our proposed method shows reasonable or better (in two scenes) results.}
    \centering
    \resizebox{0.5\textwidth}{!}{%
        \begin{tabular}{lccc}
        \toprule
        & fr1/desk & fr2/xyz & fr3/office \\
        \midrule
            NICE-SLAM\cite{nice} & 2.85 & 2.39 & 3.02\\
            ESLAM\cite{eslam}  & 2.47 & 1.11 & \textbf{2.42}\\
            \cellcolor{LightGray} LRSLAM (ours)  &
            \cellcolor{LightGray} \textbf{2.45} &
            \cellcolor{LightGray} \textbf{0.96} &
            \cellcolor{LightGray} 2.79\\
            & 
            \cellcolor{LightYellow} \red{(0.8\%$\downarrow$)} & 
            \cellcolor{LightYellow} \red{(13.5\%$\downarrow$)} & 
            \cellcolor{LightYellow} \blue{(15.3\%$\uparrow$)}\\
        \bottomrule
        \end{tabular}
    }
    \label{table:tum}
\end{wraptable}

\begin{figure*}[!t]
    \centering
    \includegraphics[width=\linewidth]{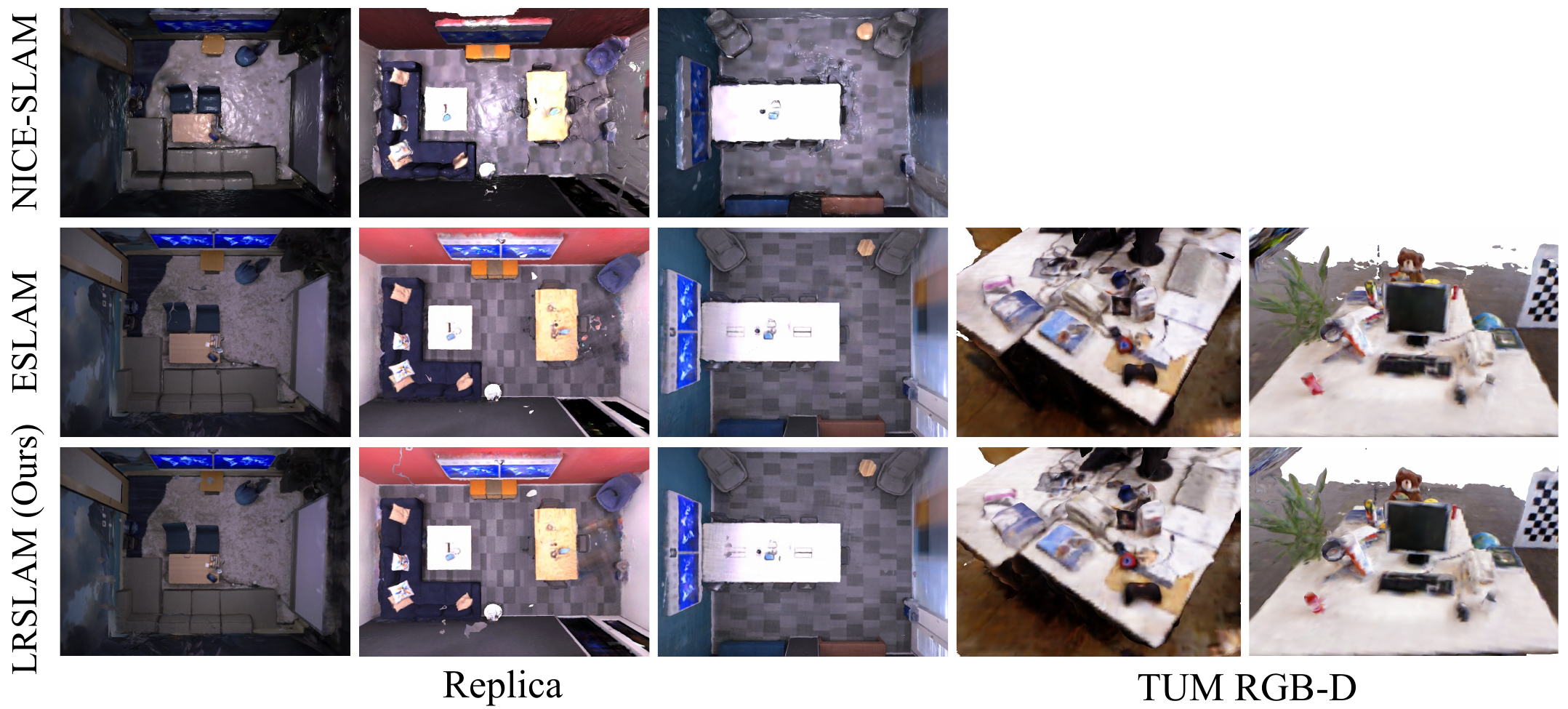}
    \caption{\textbf{Reconstruction on (a) TUM RGB-D~\cite{tum} and (b) Replica~\cite{replica}.} We qualitatively compare the quality of 3D reconstruction between our LRSLAM, NICE-SLAM~\cite{nice}, and ESLAM~\cite{eslam}. Despite using ten times more compact representation, our approach provides comparable reconstruction performance in both small-scale datasets.}
    \label{fig:replica_tum}
\end{figure*}

\myparagraph{Evaluation on TUM RGB-D~\cite{tum}.}
Further, we compare the localization accuracy on small real-world scenes from the TUM RGB-D~\cite{tum} dataset. As shown in Table~\ref{table:tum}, we compare ATE RMSE scores with existing approaches, NICE-SLAM~\cite{nice} and ESLAM~\cite{eslam}. In our experiments, all methods show reasonable reconstruction performance (i.e., ATE RMSE $\leq$ 3), where ours outperforms the others in two scenes (0.8\%--13.5\% improvements). In addition, as the dataset does not provide ground truth mesh, we provide a qualitative analysis of geometry reconstruction in Fig.~\ref{fig:replica_tum} (a). Importantly, as we reported earlier in Table~\ref{table:score_llff}, our LRSLAM requires fewer parameters (12.4\% of ESLAM), but shows comparable (or better in some scenes) geometry reconstruction quality.

{
\setlength{\tabcolsep}{2pt}
\renewcommand{\arraystretch}{1.3} 
\begin{table*}[!t]
    \caption{\textbf{Reconstruction and Localization Results on Replica~\cite{replica}.} We compare with existing approaches, including NICE-SLAM~\cite{nice} and ESLAM~\cite{eslam} regarding models' reconstruction and localization performance. The Replica dataset uses synthetic scenes with ground truth depth information, which might be unrealistic for real-world conditions. Thus, we also report scores of models with noisy depth inputs, where we add Gaussian noise $\mathcal{N}(0, 0.05^2)$ in scene \texttt{room0}. Our method shows robust performance in both reconstruction and localization.}
    %\vspace{-.5em}
        \resizebox{\linewidth}{!}{
        \begin{tabular}{lccccccc}
            \toprule
            \multirow{2}{*}{Method} & \multirow{2}{*}{\parbox{1cm}{\centering Depth Noise Added}} & &\multicolumn{2}{c}{Reconstruction Error (in cm)} && \multicolumn{2}{c}{Localization Error (in cm)} \\\cmidrule{3-8}
            & & Depth L1$\downarrow$ & Acc.$\downarrow$ & Comp.$\downarrow$ & Comp. Ratio (\%)$\uparrow$ & ATE Mean$\downarrow$ & ATE RMSE$\downarrow$\\
            \midrule
            NICE-SLAM\cite{nice} & - & 3.29 $\pm$ 0.33 & 1.66 $\pm$ 0.07 & 1.63 $\pm$ 0.05 & 96.74 $\pm$ 0.36 & 1.56 $\pm$ 0.29 & 2.05 $\pm$ 0.45 \\

            % \hline
            % NICER-SLAM\cite{nicer} & \small 3.29 $\pm$ 0.33 & \small 1.66 $\pm$ 0.07 & \small 1.63 $\pm$ 0.05 & \small 96.74 $\pm$ 0.36 & \small 1.56 $\pm$ 0.29 & \small 2.05 $\pm$ 0.45 \\
            
            ESLAM\cite{eslam} & - & 1.18 $\pm$ 0.05 & 0.97 $\pm$ 0.02 & 1.05 $\pm$ 0.01 & 98.60 $\pm$ 0.07 & 0.52 $\pm$ 0.03 & 0.63 $\pm$ 0.05 \\

            \rowcolor{LightGray} LRSLAM (ours) & - &  1.58 $\pm$ 0.11 &  1.00 $\pm$ 0.03 &  1.07 $\pm$ 0.03 & 98.94 $\pm$ 0.12 &  0.61 $\pm$ 0.04 &  0.79 $\pm$ 0.05 \\
            \midrule

            ESLAM\cite{eslam} & $\mathcal{N}(0, 0.05^2)$ &  3.22 &  2.04 &  1.99 &  98.35 &  3.00 &  2.64 \\

            \rowcolor{LightGray} LRSLAM (ours) & $\mathcal{N}(0, 0.05^2)$ &  1.20 &  1.06 & 1.18 &  98.28 & 2.94 &  2.57 \\

            & & \cellcolor{LightYellow}\red{(62.7\%$\downarrow$)} & \cellcolor{LightYellow}\red{(48.0\%$\downarrow$)} & \cellcolor{LightYellow}\red{(40.7\%$\downarrow$)} & \cellcolor{LightYellow}\blue{(0.1\%$\downarrow$)} & \cellcolor{LightYellow}\red{(2.0\%$\downarrow$)} & \cellcolor{LightYellow}\red{(2.7\%$\downarrow$)} \\
       
            \bottomrule
        \end{tabular}
        }
    %\vspace{-1ex}
    \label{table:quantitative_replica}
    % \vspace{-1.0ex}
\end{table*}
}
%WIP. Quantitative comparison of our proposed LRSLAM's reconstruction and localization accuracy with existing NeRF-based dense visual SLAM models on the Replica dataset\cite{replica}. The results are the avarage and standard deviation of five runs on eight scenes of the Replica dataset. Robustness to depth noise comparison of our method with ESLAM in term of reconstruction and Localization accuracy on the Replica dataset~\cite{replica}. Noise is added to the depth map from $\mathrm{N}(0, 0.05^2)$.

\myparagraph{Evaluation on Replica~\cite{replica}.}
We additionally compare the reconstruction and localization performance with other existing approaches, i.e., ESLAM~\cite{eslam} and NICE-SLAM~\cite{nice}, on small-scale synthetic scenes from the Replica~\cite{replica} dataset. We observe in Table~\ref{table:quantitative_replica} (see top three rows) that ESLAM and ours show reasonably good reconstruction and localization accuracy, though ours shows slightly lower scores than ESLAM, which is probably due to using a ground truth depth (i.e., synthetic) in optimizing a small-scale scene. ESLAM is more expressive than ours, tending to be overfitted to each synthetic scene easily. Thus, to make the problem more realistic, we also conduct the same experiment but with depth noise added. We observe ESLAM suffers from that noise, but ours shows robustness in both reconstruction and localization, clearly outperforming ESLAM.
From these experiments, we can reason that our low-rank representations of the scene has the ability to remove or filter the sensor noise, which is critical in RGB-D SLAM systems.

\begin{figure*}[t]
    \centering
    \includegraphics[width=\linewidth]{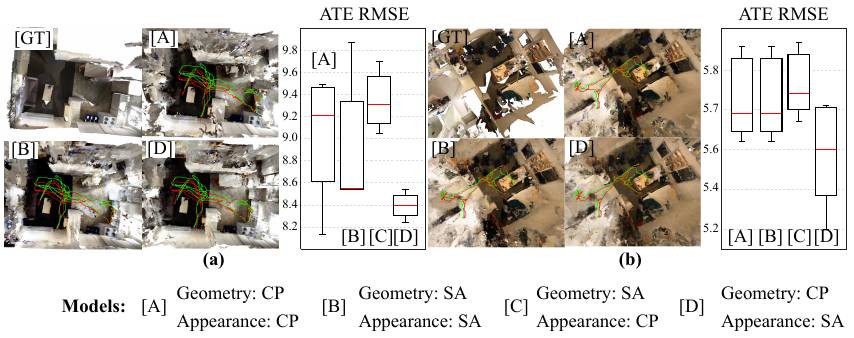}
    \caption{\textbf{Ablation Study.} Localization and reconstruction accuracy comparison between variants of our model with different combinations of scene geometry and appearance representations, i.e., CP decomposition and Six-axis (SA) decomposition. For example, the model [A] uses CP decomposition both for scene geometry and appearance representation. Note that the model [D] is ours. We visualize the results of two notable scenes from ScanNet~\cite{scannet} dataset as well as their box plots of ATE RMSE for independent five runs. More examples are provided as supplemental material.}
    \label{fig:ablation}
\end{figure*}

\subsection{Ablation Study}
\myparagraph{Comparison between Different Combinations of Scene Representation.}
In this paper, we advocate for using a combination of our proposed Six-axis decomposition and conventional compact CP decomposition, which offer an efficient memory complexity of O($n$), while the Tri-plane representation provides O($n^2$). In Fig.~\ref{fig:ablation}, we experiment to compare localization accuracy between variant models with different combinations of scene representations. In this experiment, we use large-scale real-world scenes from ScanNet~\cite{scannet} and evaluate their performance regarding ATE RMSE, the number of learnable parameters, and Frame Processing Time (FPT). As expected, Six-axis decomposition and CP decomposition clearly win in terms of representational compactness (their parameters are 4\%--6\% and 8\%--12\% less than the Tri-plane representation, respectively), mostly showing better or matched performance in localization and reconstruction accuracy. However, CP decomposition suffers from robustly reconstructing complex scenes (see variances of CP-CP model, i.e., model [A]), which is probably due to its high compactness. This may necessitate a hybrid decomposition to compensate for this trade-off relation.

\section{Conclusion}
In this paper, we presented a novel dense visual SLAM approach called LRSLAM. This approach leverages a compact scene representation based on a combination of our newly proposed Six-axis decomposition (which factorizes the three planes in the tri-plane representation into six axis-aligned feature vectors, thus holding an efficient memory complexity of O$(n)$) and conventional CP decomposition. Our experiments on three widely-used public benchmarks (i.e., ScanNet, TUM RGB-D, and Replica) validate that our proposed method indeed uses remarkably fewer parameters and shows faster processing time than existing state-of-the-art approaches, retaining matched or improved reconstruction and localization accuracy.

\subsubsection*{Acknowledgment.}
\small This work was partly supported by IITP under the Leading Generative AI Human Resources Development(IITP-2024-RS-2024-00397085, 30\%) grant, IITP grant (RS-2022-II220043, Adaptive Personality for Intelligent Agents, 30\% and IITP-2024-2020-0-01819, ICT Creative Consilience program, 10\%). This work was also partly supported by Basic Science Research Program through the NRF funded by the Ministry of Education(NRF-2021R1A6A1A13044830, 30\%).

\bibliographystyle{splncs04}
\bibliography{egbib}
\end{document}